\def\year{2021}\relax
\documentclass[letterpaper]{article} 
\usepackage{aaai21}  
\usepackage{times}  
\usepackage{helvet} 
\usepackage{courier}  
\usepackage[hyphens]{url}  
\usepackage{graphicx} 
\usepackage{natbib}  
\usepackage{caption} 
\usepackage{caption}
\usepackage{subcaption}
\usepackage{times}
\usepackage{soul}
\usepackage{graphicx}
\usepackage{amsmath}
\usepackage{booktabs}
\usepackage{enumitem}
\usepackage{amsmath}

\usepackage{subfloat}
\usepackage{amssymb,amsmath,amsthm}
\usepackage{multicol}
\usepackage{bbm}
\usepackage{microtype}
\usepackage{booktabs} 
\usepackage{algorithm}
\usepackage{algorithmic}
\usepackage{multirow}

\usepackage[switch]{lineno}  %

\newcommand{\D}{\mathcal{D}}

\newcommand{\E}{\mathbb{E}}

\newcommand{\x}{\mathbf{x}}

\newcommand{\y}{\mathbf{y}}

\newcommand{\balpha}{\boldsymbol{\alpha}}
\newcommand{\btheta}{\boldsymbol{\theta}}
\newcommand{\bbeta}{\boldsymbol{\beta}}
\newcommand{\blambda}{\boldsymbol{\lambda}}

\DeclareMathOperator*{\underE}{\mathbb{E}}

\newcommand{\calH}{\mathcal{H}}

\newtheorem{theorem}{Theorem}

\newtheorem{corollary}{Corollary}
\newtheorem{definition}{Definition}

\newcommand{\calM}{\mathcal{M}}

\title{Multi-task Learning by Leveraging the Semantic Information}
\author{

        Fan Zhou\textsuperscript{\rm 1},
        Brahim Chaib-draa\textsuperscript{\rm 1},
         Boyu Wang\textsuperscript{\rm 2,}\textsuperscript{\rm3}\thanks{The corresponding author: bwang@csd.uwo.ca (B. Wang)}\\
}
\affiliations {
    \textsuperscript{\rm 1} Université Laval, Quebec City, QC, G1V 0A6, Canada. \\
    \textsuperscript{\rm 2} University of Western Ontario, London, ON N6A 5B7, Canada \\
    \textsuperscript{\rm 3} Vector Institute, Toronto, ON M5G 1M1, Canada \\
    fan.zhou.1@ulaval.ca, chaib@ift.ulaval.ca, bwang@csd.uwo.ca}

\begin{document}
\maketitle

\begin{abstract}

One crucial objective of multi-task learning is to align distributions across tasks so that the information between them can be transferred and shared. However, existing approaches only focused on matching the marginal feature distribution while ignoring the semantic information, which may hinder the learning performance. To address this issue, we propose to leverage the label information in multi-task learning by exploring the semantic conditional relations among tasks. We first theoretically analyze the generalization bound of multi-task learning based on the notion of Jensen-Shannon divergence, which provides new insights into the value of label information in multi-task learning. Our analysis also leads to a concrete algorithm that jointly matches the semantic distribution and controls label distribution divergence. To confirm the effectiveness of the proposed method, we first compare the algorithm with several baselines on some benchmarks and then test the algorithms under label space shift conditions. Empirical results demonstrate that the proposed method could outperform most baselines and achieve state-of-the-art performance, particularly showing the benefits under the label shift conditions.
\end{abstract}
\setlength{\parindent}{0em}
\section{Introduction}
\noindent

General machine learning paradigms typically focus on learning individual tasks. Even though significant progress has been achieved, recent successes in machine learning, especially in the deep learning area, usually rely on a large amount of labelled data to obtain a small generalization error. In practice, however, acquiring labelled data could be highly prohibitive, \emph{e.g.}, when classifying multiple objects in an image \cite{long2017learning}, when analyzing patient data in healthcare data analysis~\cite{wang2015online,zhou2020task}, or when modelling users’ products preferences~\cite{murugesan2017active}. Data hungry has become a long-term problem for deep learning. Multi-task learning (MTL) aims at addressing this issue by simultaneously learning from multiple tasks and leveraging the shared knowledge across them. Many MTL approaches have been implemented in computer vision~\cite{zhao2018modulation}, natural language processing~\cite{bingel2017identifying}, medical data analysis~\cite{moeskops2016deep,li2018extracting}, brain-computer interaction~\cite{wang2020common} or cross-modality~\cite{nguyen2019multi} learning problems. It has been shown with benefits to reduce the amount of annotated data per task to reach the desired performance.

The crucial idea behind MTL is to extract and leverage the knowledge and information shared across the tasks to improve the overall performances~\cite{wang2019multitask}, which can be achieved by task-invariant feature learning ~\cite{maurer2016benefit,luo2017exploiting} or task relation learning~\cite{zhang2012convex,bingel2017identifying,zhou2020task}. One major issue with most of the existing feature learning approaches is that they only align the marginal distributions $\mathbb{P}(x)$ to extract the shared features without taking advantage of label information of the tasks. Consequently, the features can lack discriminative power for supervised learning even if their marginal features have been matched properly~\cite{dou2019domain}. Furthermore, only aligning $\mathbb{P}(x)$ cannot address the MTL problems when the label space of each task differs from each other, \emph{i.e.}, label shift problem~\cite{redko2019advances}.

While a few algorithms have been proposed to use semantic matching for MTL~\cite{zhuang2017semantic,luo2017exploiting} and have shown improved performances, the theoretical justifications for the value of labels remain elusive. Most theoretical results~\cite{ijcai2019-478,maoadaptive} for MTL derive from the notion of $\mathcal{H}$-divergence~\cite{ben2010theory} or Wasserstein adversarial training~\cite{redko2017theoretical,shen2018wasserstein}, which did not take the label information into consideration. As a result, they usually require additional assumptions, \emph{e.g.}, assuming the combined error across tasks is small~\cite{ben2010theory} to ensure the algorithms succeed, which may not hold in practice.

To this end, we propose the first theoretical analysis for MTL that considers semantic matching. Specifically, our results reveal that the MTL loss can be upper-bounded in terms of the pair-wise discrepancy between the tasks, measured by the Jensen-Shannon divergences of label distribution $\mathbb{P}(y)$ and semantic distribution $\mathbb{P}(x|y)$. 

The contributions of our work are trifold. {\bf 1.} In contrast to previous theoretical results~\cite{ijcai2019-478,maoadaptive}, which only consider the marginal distribution discrepancy (\emph{e.g.}, $\mathcal{H}$-divergence), we build a complete MTL theoretical framework upon the joint distribution discrepancy based on the Jensen-Shannon divergence. Thus, our result provides a deeper understanding of the general problem of MTL and insights into how to extract and leverage shared knowledge in a more appropriate and principled way by exploiting the label information.  {\bf 2.} Our analysis also reveals how the label shift problems affects the learning procedure of MTL, which was missing in previous results. {\bf 3.} Our theoretical result leads to a novel algorithm, namely Semantic Multi-Task Learning (SMTL) algorithm, which explicitly leverages the label information for MTL. 

Specifically, the proposed SMTL algorithm simultaneously learns task-invariant features and task similarities to match the semantic distributions across the tasks and minimizes label distribution divergence via a label re-weighting loss function. In addition, SMTL is based on a novel centroid matching approach for task-invariant feature learning, which is more efficient than other adversarial training based algorithms. To examine the effectiveness of the proposed algorithm, we evaluate SMTL on several benchmarks. The empirical results show that the proposed approach outperforms the baselines achieving state-of-the-art performance. Besides, the experiment results show that our algorithm can be more time-efficient than the adversarial baselines, which confirms the benefits of our proposed method. Furthermore, we also conduct a simulation of the label distribution shift scenario showing that the proposed algorithm could handle the label distribution shift problems that cannot be properly addressed by other baselines.

\section{Related Works}

\subsection{Multi-task learning} 

MTL aims to learn multiple tasks simultaneously and improves learning efficiency by leveraging the shared features across tasks. It has been prevalent to lots of recent machine learning topics~\cite{li2014heterogeneous,wang2016multitask,teh2017distral}.
Our work majorly relates to some feature representation learning-based approaches and task relation based approaches.~\cite{maurer2016benefit} firstly analyzed generalization error of representation-based approaches.~\cite{ murugesan2016adaptive,pentina2017multi} approached the online and transductive learning problem by MTL using a weighted summation of the losses.~\cite{wang2019transfer} analyzed the algorithmic stability in MTL. For task relations learning,~\cite{zhang2010convex,cao2018exploring} define a convex optimization problem to measure relationships while~\cite{long2017learning,kendall2018multi} propose probabilistic models by constructing task covariance matrices or estimate the multi-task likelihood via a deep Bayes model.
Latterly,~\cite{ijcai2019-478,maoadaptive} combines the feature representation learning and task relations learning together and analyzed generalization bound under the adversarial training scheme motivated by the domain adaptation problems~\cite{ben2010theory,shen2018wasserstein,shui2020beyond}, showing improved performances in vision and language processing applications, respectively. This kind of approach neglected the value of label information, which may impair the learning process. 
There were also some approaches to investigate the situation where the label space may differ from each other.  
~\cite{sutask} investigate the problem where the task samples are confusing, and the model is trained to extract task concepts by discriminating them. These sample confusing concepts also have a connection with our semantic matching approach under the label distributions shift.

\subsection{Semantic Transfer}
Leveraging semantic information has been prevalent in some machine learning topics~\cite{long2014transfer}, since it is easy to implement when few labels are available. Some few-shot adversarial based approaches~\cite{motiian2017few,luo2017label} have adopted the semantic alignment method for the learning scenario where a small amount of labelled data is available.~\cite{dou2019domain,zhou2020domain,dg_mmld} proposed to leverage the semantic information by adopting the metric learning objective under an unsupervised scheme to enforce a class-specific alignment for domain generalization problems.~\cite{zhang2019domain} propose to learn the class-specific prototype semantic information by a symmetric network to align semantic features for unsupervised domain adaptation problems.
~\cite{xie2018learning} theoretically analyzed the semantic transfer method for domain adaptation problems with a pseudo label.~\cite{shui2020beyond} investigated the value of matching conditional and semantic distribution in domain adaptation problems.
In the notion of MTL, leveraging the semantic information was investigated inconspicuously through some matrix decomposition methods in the notion of tensor learning. For example,~\cite{zhuang2017semantic} proposed a non-negative matrix factorization-based approach to learn a common semantic feature space underlying feature spaces of each
task.~\cite{luo2017exploiting} proposed to leveraging high-order statistics among tasks by analyzing the prediction weight covariance tensor of them. However, the theoretical analysis for leveraging the semantic information is still open.

\section{Notations and Preliminaries}
In this section, we start by introducing some necessary notations and preliminary problem setup.

\subsection{Problem setup}
Assuming a set of $T$ tasks $\{\hat{\D}_t\}_{t=1}^T$, each of them is generated by the underlying distribution $\D_{t}$ over $\mathcal{X}$ and by the underlying labelling functions $f_{t}:\mathcal{X}\to\mathcal{Y}$ for $\{(\D_t,f_t)\}_{t=1}^T$.
A multi-task (MTL) learner aims to find $T$ hypothesis: $h_1,\dots,h_T$ over the hypothesis space $\mathcal{H}$ to minimize the average expected error of all the tasks: 
\begin{equation*}
\arg \min_{h\in\mathcal{H}}\frac{1}{T} \sum_{i=1}^T \epsilon_t(h_t),
\end{equation*}
where $\epsilon_i(h_i) \equiv \epsilon_i(h_i,f_i) = \E_{\x\sim\D_i} \ell(h_i(\x),f_i(\x))$ is the expected error of task $t$ and $\ell$ is the loss function. For each task $i$, assume that there are $m_i$ examples. For each task $i$, we consider a minimization of weighted empirical loss for each task by defining a simplex $\balpha_j\in \Delta^T = \{\balpha_{i,j}\geq 0,~ \sum_{j=1}^T \balpha_{i,j} = 1 \}$ for the corresponding weight for task $j$. \textit{It could be viewed as an explicit indicator of the task relations revealing how much information leveraged from other tasks.}

The empirical loss $w.r.t.$ the hypothesis $h$ for task $i$ could be defined as,
\begin{equation}
\label{Eq.weighted_averge_loss}
\hat{\epsilon}_{\balpha_i}(h) = \sum_{j=1}^T \balpha_{i,j} \hat{\epsilon}_{j}(h),
\end{equation}
where $\hat{\epsilon}_{i}(h) = \frac{1}{m_i}\sum_{j=1}^{m_i} \ell(h(x_j),y_j) $ is the average empirical error for task $i$. 

Most of the existing adversarial based MTL approaches, \emph{e.g.}~\cite{maoadaptive,ijcai2019-478}, were motivated by the theory of~\cite{ben2010theory} using the $\mathcal{H}$ divergence. However, the $\mathcal{H}$-divergence theory itself is limited in many scenarios, \emph{e.g.} when tackling the (semantic) conditional shifts and understanding open set learning problems~\cite{panareda2017open,cao2018partial,you2019universal}. In this work, we adopt the \emph{Jensen-Shannon Divergence} ($D_{\text{JS}}$) to measure the differences of tasks and analyze its potentials for controlling the semantic (covariate) relations \emph{i.e.} measure the divergence between the tasks.

\begin{definition}[Jensen-Shannon divergence]
Let $\mathcal{D}_i(x,y)$ and $\mathcal{D}_j(x,y)$ be two distribution over $\mathcal{X}\times\mathcal{Y}$, and let $\calM = \frac{1}{2}(\mathcal{D}_i+\mathcal{D}_j)$, then the Jensen-Shannon (JS) divergence between $\mathcal{D}_i$ and $\mathcal{D}_j$ is,
$$D_{\text{JS}}(\mathcal{D}_i\|\mathcal{D}_j)=\frac{1}{2}[D_{\text{KL}}(\mathcal{D}_i\|\calM)+D_{\text{KL}}(\mathcal{D}_j\|\calM)]$$ 
where $D_{\text{KL}}(\mathcal{D}_i\|\mathcal{D}_j)$ is the Kullback–Leibler divergence. 
\end{definition}

It has been prevalent in adversarial training based approaches in transfer learning~\cite{dou2019domain,dg_mmld,zhao2019learning}. In practice, we could compute the \emph{Total Variation} distance ($d_{TV}$) since it is an upper bound of JS divergence~\cite{lin1991divergence}:
\begin{equation}
    d_{TV}(\mathcal{D}_i\|\mathcal{D}_j) = \frac{1}{2}|\mathcal{D}_i-\mathcal{D}_j|
\label{Eq.total_variation}
\end{equation}

\subsection{Leverage the semantic and label information}
As aforementioned, previous MTL advancements (\emph{e.g.}~\cite{maoadaptive,ijcai2019-478}) mostly only matched the marginal distribution while neglecting the labelling information. 
A successful MTL algorithm should take the semantic (covariate) conditional information into consideration. 
For example, consider the classification of different digits dataset (e.g. MNIST ($\mathcal{D}_i$) and SVHN $ (\mathcal{D}_j)$) using MTL, when conditioning on the certain digit category $Y=y$, it is clear that $\mathcal{D}_i(x|Y=y)\neq \mathcal{D}_j(x|Y=y)$, indicating the necessity of considering semantic information in MTL.

Moreover, a long-neglected issue in existing MTL approaches is that most MTL approaches all implicitly assumed that the label marginal distribution $\mathbb{P}(y)$ are the same. However, this may not hold. For example, in a medical diagnostics problem, if the data are collected from different hospitals with different populations in that area, the label spaces for data can vary from each other.~\emph{Label shift} refers to the situation where the source and target distribution have different label distribution~\cite{redko2019advances}, \emph{i.e.}, $D_{\text{JS}}(\mathcal{D}_i(y),\mathcal{D}_j(y))\neq 0$. While this issue has been investigated in literature by transfer learning~\cite{panareda2017open,geng2020recent,azizzadenesheli2019regularized}, however, the analysis towards label space shift in MTL is still open.

We show, both theoretically and empirically, that the label space shift can impair the MTL performance.  
Our theoretical and empirical results reveal that a successful multitask learning algorithm should not only match the semantic distribution $\mathcal{D}(x|y)$ among all the tasks via adversarial training with J-S divergence but also measure the label distribution $\mathcal{D}(y)$ under a re-weighting scheme for all tasks. 
Specifically, we consider the label distribution drift scenario, where the number of classes is the same to each other across the tasks while the number of instances in each class has obvious drift, \emph{i.e.}, imbalanced label distribution for all the tasks.

\section{Methodology and Theoretical Insights}

Intuitively, when aligning the distribution of different tasks, features from the same class should be mapped near to each other in the feature space satisfying the semantic conditional relations. 
We firstly analyze the error bound with the notion of Jensen-Shannon divergence based form to measure the tasks discrepancies. Then, we further extend the results to analyze to control the label space divergence and the semantic conditional distribution divergence. 
\emph{All the proofs are delegated to the supplementary materials.}
\begin{theorem}\label{aaai_theorem1}
Let $\mathcal{H}$ be the hypothesis class $h\in\mathcal{H}$. Suppose we have $T$ tasks generated by the underlying distribution and labelling function $\{(\D_1,f_1),\dots, (\D_T,f_T)\}$. Assume the loss function $\ell$ is bounded by $L$ ($\max(\ell)-\min(\ell)\leq L$). Then, with high probability we have
\begin{equation*}
\begin{split}
    \frac{1}{T}\sum_{t=1}^T\epsilon_t(h)&\leq\frac{1}{T}\sum_{t=1}^T\epsilon_{\alpha_t}(h)+ \frac{\lambda_0}{4T}L^2\\
    &+ \frac{2}{\lambda_0 T} \sum_{t=1}^T \sum_{i=1}^T\balpha_{t,i}D_{\text{JS}}(\mathcal{D}_t(x,y)\|\mathcal{D}_i(x,y))
\end{split}
\end{equation*}
where $\lambda_0> 0$ is a constant.
\end{theorem}

Theorem~\ref{aaai_theorem1} showed that the averaged MTL error is bounded by an averaged summation of all the tasks, the averaged summation of task distribution divergence among all pair of tasks and some constant value.

This bound indicates the joint distribution while we aim to leverage the label ($\mathbb{P}(y)$) and semantic ($\mathbb{P}(x|y)$) information, based on the aforementioned theorem 1, we could then decompose it into the following results,

\begin{corollary}
\label{corollary_label_semantic}
Follow the setting of Theorem~\ref{aaai_theorem1}, we can further bound the overall task error by 
\begin{equation*}
        \begin{split}
    \frac{1}{T}\sum_{t=1}^T\epsilon_t(h)&\leq\frac{1}{T}\sum_{t=1}^T\epsilon_{\alpha_t}(h) + \blambda \underbrace{ D_{\text{JS}}(\mathcal{D}_t(y)\|\mathcal{D}_i(y))}_{\text{Label distribution divergence}} \\
    &+ \blambda \underbrace{\underE_{y\sim\mathcal{D}_t(y)}D_{\text{JS}}(\mathcal{D}_t(x|y)\|\mathcal{D}_i(x|y))}_{\text{Semantic distribution divergence}} \\
    & + \blambda  \underbrace{\underE_{y\sim\mathcal{D}_i(y)} D_{\text{JS}}(\mathcal{D}_t(x|y)\|\mathcal{D}_i(x|y))}_{\text{Semantic distribution divergence}} +\frac{\lambda_0}{4T}L^2 
    \end{split}
    \end{equation*}
    where $\blambda\in \mathbb{R}^{T\times T}$ is the corresponding matrix whose $t$-th row and $i$-th column element is $\frac{2}{\lambda_0 T} \sum_{t=1}^T \sum_{i=1}^T\balpha_{t,i}$
\end{corollary}
\paragraph{Remark:}Different from~\cite{ijcai2019-478,maoadaptive}, which were motivated by~\cite{ben2010theory}, our theoretical results do not rely on extra assumption of the existence of the optimal hypothesis to achieve a small combined error. Besides, our results also provide new insight by take advantage of label information and semantic conditional relations.

Corollary~\ref{corollary_label_semantic} implies that the averaged error over all tasks is bounded by the summation of task errors (the first term in \emph{R.H.S.} of Corollary~\ref{corollary_label_semantic}), the label distribution divergence (the second term), a constant term (the third term), and the semantic distribution divergence term (the last two terms). The first term could be easily optimized by a general supervised learning loss (\emph{e.g.} the cross-entropy loss). To minimize this bound now is equivalent to match the semantic distribution among the tasks and measure the label divergence. Since the labels of each task samples are available to the learner, we could leverage the label and semantic information directly.

\begin{algorithm}[tb]
\caption{The Global Semantic Matching Method}
\label{alg:semantic_transfer}
\textbf{Input}: Training set from each tasks\\
\textbf{Parameter}: Feature extractor $\btheta^f$; decay parameter $\gamma$ \\
\textbf{Output}: The semantic loss
\begin{algorithmic}[1] 
\FOR{k=1 to K}
\STATE $C_{\mathcal{D}_i}^k \leftarrow \frac{1}{|\mathcal{D}_i^k|}\sum_{(x_i,y_i)\in\mathcal{D}_i^k}\btheta^f(x_i)$ 
\STATE $C_{\mathcal{D}_j}^k \leftarrow \frac{1}{|\mathcal{D}_j^k|}\sum_{(x_j,y_j)\in\mathcal{D}_j^k}\btheta^f(x_j)$
\STATE $\mathbf{C}_{\mathcal{D}_i}^k \leftarrow \gamma C_{\mathcal{D}_i}^k + (1-\gamma)C_{\mathcal{D}_i}^k $
\STATE $\mathbf{C}_{\mathcal{D}_j}^k \leftarrow \gamma C_{\mathcal{D}_j}^k + (1-\gamma)C_{\mathcal{D}_j}^k $
\STATE $\mathcal{L}_S \leftarrow \mathcal{L}_S +\Phi(C_{\mathcal{D}_i},C_{\mathcal{D}_j})$
\ENDFOR
\STATE \textbf{return} $\mathcal{L}_S$
\end{algorithmic}
\end{algorithm}


\begin{table*}[t]
    \centering
    \resizebox{1\textwidth}{!}{
\Huge\begin{tabular}{l|ccc|c|ccc|c|ccc|c}
\hline
                    & \multicolumn{4}{c|}{\textbf{3K}}                                                        & \multicolumn{4}{c|}{\textbf{5K}}                                                                  & \multicolumn{4}{c}{\textbf{8K}}                                                                        \\
\textbf{Approach}   & \textbf{MNIST}        & \textbf{MNIST-M}      & \textbf{SVHN}        & \textbf{Avg.} & \textbf{MNIST}        & \textbf{MNIST-M}       & \textbf{SVHN}         & \textbf{Avg.}         & \textbf{MNIST}        & \textbf{MNIST-M}      & \textbf{SVHN}               & \textbf{Avg.}          \\ \hline
\textbf{Vanilla}    & $93.9\pm 3.2$         & $77.1\pm2.6$          & $57.3\pm 0.4$        & $76.1$           & $96.3\pm1.2$          & $79.1\pm3.1$           & $68.0\pm2.9$          & $81.1$                     & $97.7\pm 0.5$         & $83.7\pm 2.2$         & $71.4\pm0.9$                & $84.2$                    \\
\textbf{Weighed}   & $89.3\pm3.3$          & $76.4\pm3.1$          & $70.2\pm1.8$         & $78.3$           & $91.8\pm2.7$          & $74.2\pm0.9$           & $73.6\pm3.1$          & $79.8$                   & $92.3\pm2.6$          & $76.9\pm3.1$          & $74.1\pm1.6$                & $81.1$                    \\
\textbf{Adv.H}      & $90.1\pm1.2$          & $\mathbf{81.2}\pm1.3$ & $70.8\pm0.5$         & $80.7$           & $91.9\pm2.6$          & $\mathbf{83.7}\pm1.4$           & $73.6\pm1.6$          & $82.9$                     & $94.9\pm1.6$          & $\mathbf{85.2}\pm0.3$          & $79.1\pm0.3$                & $86.4$                    \\
\textbf{Adv.W}      & $96.8\pm0.6$          & $81.3\pm0.7$          & $69.5\pm1.1$         & $82.5$           & $97.5\pm0.2$          & $83.4\pm0.4$             & $72.6\pm1.2$            & $84.5$                     & $98.1\pm0.3$          & $84.3\pm0.4$          & $75.4\pm1.1$ & $86.1$                    \\
\textbf{Multi-Obj.} & $\mathbf{97.5}\pm0.3$ & $76.9\pm0.5$          & $54.8\pm0.3$         & $76.4$           & $\mathbf{98.2}\pm0.2$ & $80.2\pm0.7$           & $61.2\pm0.8$          & $79.9$                     & $\mathbf{98.5}\pm0.3$ & $82.8\pm0.5$          & $69.9\pm0.9$                & $83.7$                    \\
\textbf{AMTNN}      & $96.9\pm0.2$         & $80.8\pm1.5$          & $77.1\pm0.9$         & $84.9$           & $97.7\pm 0.1$         & $83.6\pm 1.1$ & $78.4\pm0.8$          & $86.6$                   & $98.1\pm0.2$          & $83.1\pm2.1$          & $80.2\pm1.3$                & $87.1$                    \\ \hline
\textbf{Ours}       & $95.4\pm0.3$          & $80.1\pm0.5$          & $\mathbf{81.5}\pm0.6$ & $\mathbf{85.7}$  & $95.8\pm 0.3$        & $82.4\pm 0.4$          & $\mathbf{83.3}\pm0.3$ & $\mathbf{87.2}$ & $96.0\pm0.3$         & $83.9\pm0.4$ & $\mathbf{85.4}\pm0.2$       & \textbf{$\mathbf{88.4}$} \\ \hline
\end{tabular}
}
    \caption{The empirical results (in $\%$) on the digits datasets.}
    \label{tab:digit}
\end{table*}


\subsection{Label Re-weighting Loss}

Corollary~\ref{corollary_label_semantic} indicates that the error is also controlled by the label divergence term $D_{\text{JS}}(\mathcal{D}_i(y)\|\mathcal{D}_j(y))$. To reduce the influences caused by the label space shifts, we could adopt a label correction re-weighting loss~\cite{lipton2018detecting} based on the number of instances in each class,

\begin{equation}
    \hat{\epsilon}^{\beta}_{\mathcal{D}_i}(h) = \sum_{(x_i,y_i)\in\hat{\mathcal{D}_i}} \bbeta(y_i) \ell(h(x_i),y_i))
\end{equation}
where $\bbeta\in\mathbb{R}^{K\times1}$ is weight for each class, and $\bbeta_j$ is the weight for class $y_j$. For task $i$ with total $m_i$ instances, the weight of class $k\in K$ ($K$ is the total number of classes) is computed by $\bbeta_k =\sum \frac{|\# y = y_k |}{m_i}$. This re-weighting scheme guarantees the instances from different classes could have equal probability to be sampled when training the model, which re-weights the loss according to frequency of each class that occurs during training. By doing so, the learner will not neglect those classes who have fewer instances and therefore takes care of label drift.

 \textbf{Note}: the coefficient $\bbeta$ is computed for re-weighting the loss from each task while $\balpha$ is a set of weights indicating the relations between each other, \emph{i.e.} how much information leveraged from other tasks.
 
When training the model, we maintain the task specific loss $\mathcal{L}_{i} = \hat{\epsilon}^{\beta}_{\mathcal{D}_i}(h) $ and compute the total classification loss
\begin{equation}
\label{classificaiton_obj}
    \mathcal{L}_C = \sum_{i=1}^T \balpha_{i}\hat{\epsilon}^{\beta}_{\mathcal{D}_i}(h)
\end{equation}

\subsection{Semantic matching and task relation update}
 To compute Eq.~\ref{classificaiton_obj}, we still need to estimate the task relation coefficients $\balpha$. As it indicates the relations between tasks, we are not able to measure its value at the beginning. To a better estimation, we need to update the coefficient $\balpha$ automatically during the training process. Through Corollary~\ref{corollary_label_semantic}, we could solve the coefficients via an convex optimization as
 
 \begin{equation}
 \label{convex_opt_alpha}
    \begin{split}
          \min_{\balpha_1,\dots,\balpha_T} & \quad \mathcal{L}_C +\sum_{i,t=1}^T\balpha_{t,i}  \sum\nolimits_{y} (\hat{\mathcal{D}}_i(y) +\hat{\mathcal{D}}_t(y))) \mathbf{E}_{i,t}   \\
          & +\sum_{t=1}^T||\balpha_t||_2\\
         & \text{s.t.}\quad \sum_{t} \balpha_t = 1
    \end{split}
\end{equation}
where $\mathbf{E}_{i,t} = D_{\text{JS}}\left(\hat{\mathcal{D}_t}(x|y)\|\hat{\mathcal{D}_t}(x|y)\right)$ is the empirical semantic distribution divergence. 

 To align the semantic distribution, we adopt the centroid matching method by computing the Euclidean distance between two centroids in the embedding space. Denote $C_{\mathcal{D}_i}^k$ and $C_{\mathcal{D}_j}^k$ by two feature centroids from class $k$ of distribution $\mathcal{D}_i$ and $\mathcal{D}_j$ respectively, it could be computed by,
 \begin{equation}
     \Phi(C_{\mathcal{D}_i}^k,C_{\mathcal{D}_j}^k) = ||C_{\mathcal{D}_i}^k -C_{\mathcal{D}_j}^k ||^2
     \label{centroid_computing}
 \end{equation}

\begin{figure}
    \centering
    \includegraphics[width=1\linewidth]{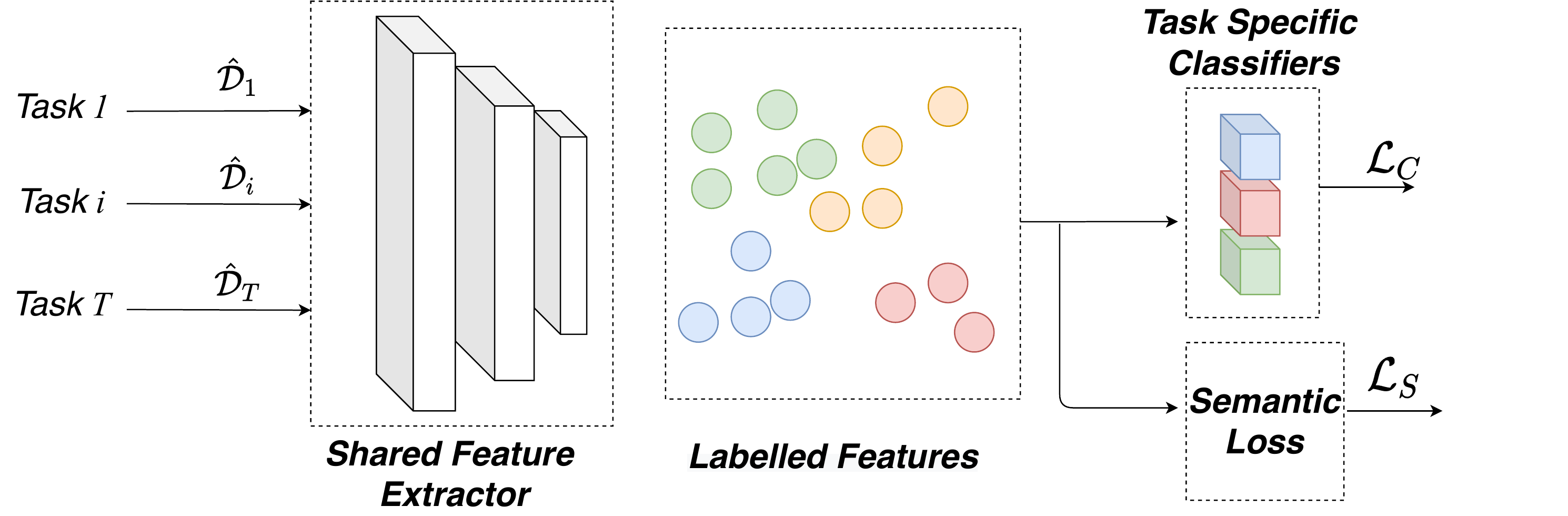}
    \caption{The overall model architecture}
    \label{The_overall_model_architecture}
\end{figure}

\begin{table*}[]
\centering
 \resizebox{1.0\textwidth}{!}{\Huge \begin{tabular}{l|ccccc|ccccc|ccccc}
\hline
\textbf{}           & \multicolumn{5}{c|}{\textbf{$10\%$}}                              & \multicolumn{5}{c|}{\textbf{$15\%$}}                              & \multicolumn{5}{c}{\textbf{$20\%$}}                              \\
\textbf{Approach}   & \textbf{A} & \textbf{C} & \textbf{P} & \textbf{S} & \textbf{avg.} & \textbf{A} & \textbf{C} & \textbf{P} & \textbf{S} & \textbf{avg.} & \textbf{A} & \textbf{C} & \textbf{P} & \textbf{S} & \textbf{avg.} \\ \hline
\textbf{Vanilla MTL}      & $78.8\pm1.1$      & $81.8\pm1.4$      & $87.1\pm0.4$      & $83.4\pm0.6$      & $82.8$         & $82.8\pm0.8$      & $86.7\pm0.9$      & $89.3\pm0.7$      & $84.9\pm0.7$      & $85.9$         & $84.1\pm0.9$      & $87.9\pm0.8$      & $90.5\pm0.6$      & $85.8\pm1.2$      & $87.1$              \\
\textbf{Weighted}   &  $82.7\pm1.1$          & $86.2\pm0.6$      &  $89.7\pm1.1$  & $84.9\pm0.7$  &    $85.9$      &    $\mathbf{85.1}\pm0.9$        &  $87.9\pm0.7$  & $91.2\pm1.0$           &  $87.3\pm0.7$        & $87.9$              &   $86.1\pm0.6$         &  $89.7\pm0.4$  & $92.1\pm1.2$  & $88.4\pm0.9$           &  $89.1$             \\
\textbf{Adv.W}      & $78.8\pm1.1$      & $83.9\pm1.2$      & $87.6\pm1.4$      & $84.0\pm0.9$      & $83.6$         & $83.6\pm1.4$      & $84.8\pm0.5$      & $84.7\pm0.7$      & $84.0\pm0.6$      & $84.2$         & $83.5\pm0.5$      & $89.5\pm0.7$      & $91.4\pm0.5$      & $87.3\pm0.6$      & $87.9$         \\
\textbf{Adv.H}      & $76.8\pm1.6$      & $84.3\pm0.4$      & $88.3\pm0.6$      & $84.0\pm0.6$      & $83.3$         & $82.6\pm0.8$      & $87.8\pm0.7$      & $89.9\pm0.7$      & $86.1\pm0.4$      & $86.6$         & $84.4\pm0.7$      & $87.6\pm0.3$      & $91.5\pm0.4$      & $88.3\pm0.4$      & $87.9$         \\
\textbf{Multi-Obj.} & $79.4\pm1.8$      & $83.4\pm1.3$      & $87.0\pm0.5$      & $82.9\pm1.1$      & $83.2$         & $82.7\pm0.5$      & $87.5\pm0.4$      & $89.1\pm0.6$      & $86.5\pm0.5$      & $86.4$         & $84.3\pm0.9$      & $88.7\pm0.4$      & $91.0\pm0.3$      & $88.8\pm0.7$      & $88.2$         \\
\textbf{AMTNN}      & $\mathbf{82.8}\pm0.4$      & $86.7\pm 0.4$      & $91.3\pm0.9$      & $81.2\pm0.8$      & $85.5$         & $85.1\pm0.3$      & $88.8\pm0.3$      & $92.9\pm0.4$      & $85.8\pm0.5$      & ${88.2}$         & $\mathbf{87.4}\pm0.2$      & $89.9\pm0.6$      & $93.7\pm0.4$      & $87.7\pm0.1$       & ${89.7}$         \\ \hline
\textbf{Ours}   & $80.6\pm 0.4$   & $\mathbf{87.9}\pm0.4$  & $\mathbf{94.4}\pm 0.5$ & $\mathbf{91.9}\pm 0.5$   &  $\mathbf{88.9}$             &  $83.4\pm0.4$          & $\mathbf{89.8}\pm0.6$        &    $\mathbf{94.5}\pm0.4$  &  $\mathbf{93.1}\pm 0.4$          &  $\mathbf{90.2}$   & $86.3\pm0.3$  & $\mathbf{91.6}\pm0.4$ & $\mathbf{95.5}\pm0.4$           & $\mathbf{93.8}\pm 0.4$           &  $\mathbf{91.8}$             \\ \hline
\end{tabular}}
\caption{The empirical results (in \%) on PACS dataset with AlexNet as feature extractor.}
\label{Table_results_PACS}
\end{table*}


 Our goal is to match the semantic distribution across tasks. For this, we re-visited the \emph{moving average centroid} method by~\cite{xie2018learning} where a global centroid matrix was maintained to compute the semantic information between a labeled source and an unlabeled target distribution for domain adaptation problem. Unlike~\cite{xie2018learning}, we could explicitly measure the semantic distribution across all the tasks rather than through assigning pseudo labels to compute them. We illustrate the modified moving average centroid method, namely~\emph{The Global Semantic Matching Method} in Algorithm.~\ref{alg:semantic_transfer}. 
 \paragraph{Remark:} Through Algorithm~\ref{alg:semantic_transfer}, the semantic loss $\mathcal{L}_S$ is an approximation of the \emph{total variation distance} (see Eq.~\ref{Eq.total_variation}) of the two centroids, which is an upper bound of $\mathcal{D}_{\text{JS}}\left(\mathcal{D}_i(x|y),\mathcal{D}_j(x|y) \right)$. Compared with adversarial training based method, this semantic matching process does not need to train pair-wised discriminators, which may help to reduce the computational costs. For example, for $m$ tasks, ~\cite{ijcai2019-478} needs to train $\frac{m(m-1)}{2}$ discriminators. When the number of tasks increase, the training procedure may become time-inefficient.

\begin{center}
\begin{algorithm}[]
		\caption{The proposed Semantic Multi-task learning algorithm}
		\begin{algorithmic}[1] 
		\REQUIRE Samples from different tasks $\{\hat{\D}_t\}_{t=1}^T$, initial coefficients $\{\balpha_t\}_{t=1}^T$ and learning rate $\eta$
        \ENSURE Neural network $\btheta^{f}$, $\{\btheta_{t}^C\}_{t=1}^T$ and coefficient $\balpha_1,\dots,\balpha_T$
        \WHILE{Algorithm Not converge}
        \FOR{min-batch $\{(\x^b_t,\y^b_t)\}$ from task $\{\hat{\D}_t\}_{t=1}^T$}
        \STATE Compute the classification objective $\mathcal{L}_C$ by Eq.~\ref{classificaiton_obj}
        \STATE Compute the semantic matching objective $\mathcal{L}_S$ via Algorithm~\ref{alg:semantic_transfer}
        \STATE Update the network parameters $\btheta^f,\btheta^c_{t}$ by: \\$\btheta^f \leftarrow \btheta^f -\eta  \frac{\partial \mathcal{L}_C + \mathcal{L}_S}{\partial\btheta^f}$ and  $\btheta^c_{t}\leftarrow \btheta^c_{t} - \eta \frac{\partial \mathcal{L}_C + \mathcal{L}_S}{\partial\btheta^c_{t}}$ 
        \ENDFOR
		\STATE Update $\{\balpha_t\}_{t=1}^T$ by optimizing over Eq.~(\ref{convex_opt_alpha}).
		\ENDWHILE
        \end{algorithmic}
        \label{asmtnn_algo}
\end{algorithm}
\end{center}





\subsection{The full objective and proposed algorithm}
With the key components introduced in previous sections, we could summarize the full method. A general model architecture is provided in Fig.~\ref{The_overall_model_architecture}. The model learns multiple tasks jointly by a shared feature extractor. For each task, we implement a task-specific classifier. The classifier was trained under a re-weighting loss via measuring label distribution of each task, and we also maintain the semantic loss to match the semantic distribution across tasks to achieve the semantic transfer objective. The proposed Semantic Multi-task learning (SMTL) method is illustrated in Algorithm~\ref{asmtnn_algo}


\section{Experiments and Analysis}
\begin{table}[]
{
\centering
\resizebox{\linewidth}{!}{\Large\begin{tabular}{@{}l|llll|l@{}}
\toprule
\textbf{Method}         & \textbf{Amazon}         & \textbf{Caltech}        & \textbf{Dslr}          & \textbf{WebCam}         & \textbf{Average}        \\ \midrule
\textbf{Vanilla}    & $84.2\pm1.1$          & $80.6\pm0.8$          & $90.8\pm2.3$          & $81.8\pm0.9$          & $84.3$   \\
\textbf{Weighted}       & $88.1\pm0.2$          & $81.5\pm0.9$          & $94.9\pm0.2$          & $94.2\pm0.5$          & $88.6$          \\
Adv.H         & $81.5\pm0.5$          & $73.8\pm1.8$          & $91.4\pm2.1$          & $86.1\pm 1.4$          & $83.3$          \\
Adv.W         & $84.9\pm0.4$          & $80.9\pm0.9$          & $94.5\pm2.2$          & $87.5\pm1.5$          & $86.9$          \\
Multi-Obj.    & $82.3\pm0.7$            & $76.7\pm2.4$        &     $91.2\pm1.7$      & $86.8\pm0.9$          &     $84.3$      \\
AMTNN         & $89.3\pm 0.9$          & $84.3\pm0.6$          & $\mathbf{98.4}\pm1.3$          & $94.1\pm0.7$          & $91.7$          \\ \midrule 
Ours          & $\mathbf{90.9}\pm0.4$             & $\mathbf{85.3}\pm0.5$            & $98.1\pm0.8$           &$\mathbf{94.2}\pm0.6$          &  $\mathbf{92.1}$           \\ \bottomrule
\end{tabular}}}
\caption{Average test accuracy (in \%) of MTL algorithms on Office-Caltech dataset with AlexNet as feature extractor.}
\label{caltech}
\end{table}

\begin{table*}[]
\centering
\resizebox{\linewidth}{!}{\large\begin{tabular}{@{}l|cccc|cccc|cccc@{}}
\toprule
\textbf{}         & \multicolumn{4}{c|}{\textbf{$5\%$}}                               & \multicolumn{4}{c|}{\textbf{$10\%$}}                              & \multicolumn{4}{c}{\textbf{$20\%$}}                              \\
\textbf{Approach} & \textbf{Amazon} & \textbf{Dslr} & \textbf{Webcam} & \textbf{avg.} & \textbf{Amazon} & \textbf{Dslr} & \textbf{Webcam} & \textbf{avg.} & \textbf{Amazon} & \textbf{Dslr} & \textbf{Webcam} & \textbf{avg.} \\ \midrule
\textbf{Vanilla}     & $61.3\pm1.3$           & $71.8\pm2.1$         & $72.1\pm1.1$           & $68.3$         &   $73.2\pm0.5$    &  $80.6\pm1.4$             &    $82.1\pm0.9$     &  $78.6$             & $79.4\pm0.8$           & $91.2\pm1.0$         & $93.1\pm0.8$           & $87.9$         \\
\textbf{Weighted} &  $63.3\pm0.2$       & $87.4\pm2.3$      & $84.9\pm0.6$   & $78.5$   &  $70.6\pm1.2$     &       $92.1\pm0.9$    &  $88.4\pm1.3$      &  $83.7$ &     $76.8\pm0.9$            &    $\mathbf{96.6}\pm0.7$           &      $95.6\pm0.5$           &    $89.7$           \\
\textbf{Adv.W}  & $66.5\pm1.9$           & $71.8\pm1.1$         & $69.9\pm0.9$           & $69.7$         & $74.7\pm1.1$           & $85.9\pm0.8$         & $85.7\pm0.8$           & $82.1$        & $79.3\pm0.6$           & $93.8\pm0.4$         & $92.2\pm0.9$           & $88.4$         \\
\textbf{Adv.H}  & $65.8\pm1.1$           & $73.5\pm0.8$         & $71.4\pm0.7$           & $70.2$         & $71.0\pm0.9$           & $84.1\pm0.9$         & $89.4\pm0.1$           & $81.4$         & $79.7\pm0.5$           & $93.7\pm0.7$         & $93.7\pm0.6$           & $89.1$         \\
\textbf{Multi-Obj.}      & $\mathbf{68.9}\pm1.2$           & $72.5\pm1.4$         & $72.3\pm0.4$           & $71.3$         & $74.6\pm0.9$           & $86.8\pm1.1$         & $86.9\pm0.8$           & $82.8$         & $79.2\pm0.8$           & $92.1\pm0.6$         & $94.7\pm0.6$           & $88.6$         \\
\textbf{AMTNN}    &  $63.3\pm0.6$    & $80.1\pm1.6$     &  $85.4\pm0.3$ & $79.3$ & $71.3\pm1.2$        &    $\mathbf{92.8}\pm0.9$  & $89.6\pm1.2$    &  $84.6$            &  $80.2\pm0.9$  &    $94.2\pm1.2$   &$94.4\pm0.9$   &   $89.6$             \\ \midrule
\textbf{Ours}     & $68.5\pm0.6$           & $\mathbf{87.9}\pm0.8$         & $\mathbf{86.5}\pm0.5$           & $\mathbf{80.9}$         & $\mathbf{75.7}\pm0.2$           & $\mathbf{92.8}\pm0.2$         & $\mathbf{90.8}\pm0.3$           & $\mathbf{86.4}$         & $\mathbf{81.1}\pm0.2$          & $96.5\pm0.1$         & $\mathbf{96.1}\pm0.2$           & $\mathbf{91.2}$         \\ \bottomrule
\end{tabular}}
\caption{The empirical results (in \%) on Office-31 dataset with ResNet-18 as feature extractor.}
\label{Results_office-31}
\end{table*}

In order to investigate the effectiveness of our method, we examined the proposed approach comparing with several baselines on \textbf{Digits}, \textbf{PACS}~\cite{li2017deeper}, \textbf{Office-31}~\cite{saenko2010adapting}, \textbf{Office-Caltech}~\cite{gong2012geodesic} and \textbf{Office-home}~\cite{venkateswara2017Deep} dataset. For the Digits benchmark, we evaluate the algorithms on \emph{MNIST}, \emph{MNIST-M} and \emph{SVHN} simultaneously. The PACS dataset, which was widely used in recent transfer learning researches, consists of images from four tasks: \emph{Photo} (P), \emph{Art painting} (A), \emph{Cartoon} (C), Sketch (S), with objects from 7 classes. Office-31 dataset is a vision benchmark widely used in transfer learning related problems which consists of three different tasks: \emph{Amazon}, \emph{Dslr} and \emph{Webcam}; Office-Caltech contains the 10 shared categories between the Office-31 dataset and Caltech256 dataset, including four different tasks: \emph{Amazon}, \emph{Dslr}, \emph{Webcam} and \emph{Caltech}; Office-home is a more challenging benchmark, which contains four different tasks: \emph{Art}, \emph{Clipart}, \emph{Product} and \emph{Real World}, with $65$ categories in each task.

\begin{table*}[]
\centering
\resizebox{\linewidth}{!}{
\Huge
\begin{tabular}{@{}lccccc|ccccc|ccccc@{}}
\toprule
\textbf{}                              & \multicolumn{5}{c}{\textbf{$5\%$}}                                                       & \multicolumn{5}{c|}{\textbf{$10\%$}}                                                     & \multicolumn{5}{c}{\textbf{$20\%$}}                                                     \\ \midrule
\multicolumn{1}{l|}{\textbf{Approach}} & \textbf{Art} & \textbf{Clipart} & \textbf{Product} & \textbf{Real-world} & \textbf{avg.} & \textbf{Art} & \textbf{Clipart} & \textbf{Product} & \textbf{Real-world} & \textbf{avg.} & \textbf{Art} & \textbf{Clipart} & \textbf{Product} & \textbf{Real-world} & \textbf{avg.} \\ \midrule
\multicolumn{1}{l|}{\textbf{Vanilla}}     & $26.2\pm0.3$                  & $30.1\pm0.2$                      & $57.6\pm0.1$                      & $47.4\pm1.1$                         & 40.3                           & $35.8\pm0.7$                                       & $43.3\pm0.6$                        & $67.1\pm0.4$                      & $56.8\pm1.3$                         & $50.7$              &      $45.5\pm0.8$        &       $56.1\pm0.6$       &        $74.4\pm0.7$  &     $62.6\pm0.6$        & $59.6$      \\
\multicolumn{1}{l|}{\textbf{Weighted}} &        $26.8\pm1.6$      &     $31.8\pm1.8$       &    $59.2\pm0.4$      &     $50.5\pm1.2$    & $42.1$     & $38.2\pm1.0$     &   $45.3\pm1.6$  &     $69.1\pm0.2$    & $58.3\pm0.8$        & $52.7$    &   $47.9\pm0.1$   &  $56.7\pm0.9$                & $75.6\pm0.6$   &  $64.8\pm0.9$                   &    $61.2$           \\
\multicolumn{1}{l|}{\textbf{Adv.W}}  & $26.8\pm0.8$     & $32.7\pm0.5$      &      $58.3\pm0.9$            &   $47.1\pm0.4$      &    $41.2$    &       $38.5\pm0.8$   &    $44.4\pm0.7$              & $67.6\pm0.7$                 & $59.5\pm0.9$                  & $52.3$         &     $47.9\pm0.5$      &    $56.7\pm0.6$       &    $75.4\pm1.1$    & $65.7\pm0.8$    &   $61.3$            \\
\multicolumn{1}{l|}{\textbf{Adv.H}} & $27.7\pm1.4$ & $32.1\pm1.5$                      & $59.6\pm0.7$                      & $51.1\pm0.9$                         & $42.7$ &      $39.0\pm0.9$   & $45.8\pm1.8$  &     $69.4\pm0.4$     &          $58.8\pm0.6$  &     $53.2$          &      $46.7\pm0.5$        & $56.5\pm1.1$  & $75.6\pm0.4$              & $65.1\pm0.7$                     & $61.0$               \\
\multicolumn{1}{l|}{\textbf{Multi-Obj.}}      &   $25.6\pm1.5$           &      $31.7\pm1.7$            & $58.7\pm1.3$                 &    $51.5\pm0.9$                 &      $41.8$         &    $34.6\pm0.9$          & $43.3\pm1.4$    & $66.1\pm1.5$                 &    $56.8\pm0.7$ &  $50.2$      & $46.2\pm0.8$       & $56.6\pm0.5$     &  $74.3\pm0.7$                & $62.8\pm0.6$  &     $59.8$                               \\
\multicolumn{1}{l|}{\textbf{AMTNN}}    & $32.5\pm1.3$   & $34.5\pm0.9$       & $56.3\pm0.8$     & $49.9\pm1.8$        & $43.3$  & $41.1\pm1.0$ &    $47.5\pm0.8$ & $68.4\pm0.7$     & $58.9\pm0.9$       &  $53.9$  &  $48.9\pm0.5$     & $\mathbf{60.7}\pm0.4$          & $75.4\pm0.4$    & $64.7\pm0.4$ & $62.1$             \\ \midrule
\multicolumn{1}{l|}{\textbf{Ours}}      & $\mathbf{38.3}\pm0.9$ & $\mathbf{40.9}\pm0.9$ & $\mathbf{62.3}\pm0.8$ & $\mathbf{55.5}\pm0.6$ & $\mathbf{49.2}$                         & $\mathbf{43.8}\pm0.6$                  & $\mathbf{50.4}\pm0.8$                      & $\mathbf{71.3}\pm0.9$         & $\mathbf{62.3}\pm0.6$                        & $\mathbf{57.1}$                         & $\mathbf{51.2}\pm0.7$                  & $60.6\pm0.8$                      & $\mathbf{77.9}\pm0.4$                      & $\mathbf{66.1}\pm0.6$                         & $\mathbf{64.3}$            \\ \bottomrule
\end{tabular}}
\caption{The empirical results (in \%) on Office-home dataset with ResNet-18 as feature extractor. }
\label{results_office-home}
\end{table*}

\begin{table}[]
{
\centering
\resizebox{\linewidth}{!}{\begin{tabular}{@{}l|lll|l@{}}
\toprule
\textbf{Method}         & \textbf{Amazon}               & \textbf{Dslr}          & \textbf{WebCam}         & \textbf{Average}        \\ \midrule
Cls. only  & $79.4\pm0.8$           & $91.2\pm1.0$         & $93.1\pm0.8$           & $87.9$        \\
$w.o.$ re-weighting     & $80.2\pm0.7$ & $94.7\pm1.3$                & $94.1\pm0.8$          & $89.6$          \\
$w.o.$ sem. matching    & $79.8\pm0.5$          & $96.1\pm0.3$          & $95.4\pm0.3$          & $90.4$          \\
$w.o.$ cvx opt.         & $80.7\pm 0.3$      & $\mathbf{96.8}\pm0.5$         & $95.3\pm0.4$          & $90.9$          \\
Full method           &   $81.1\pm0.2$          & $96.5\pm0.1$         & $96.1\pm0.2$           & $\mathbf{91.2}$                \\ \bottomrule
\end{tabular}}}
\caption{Ablation studies on Office-31 dataset.}
\label{ablation_studies}
\end{table}

To evaluate the performance of our proposed algorithm, we re-implement and compare our method with the following principled approaches:
\begin{itemize}
    \item Vanilla MTL: Learning all the tasks simultaneously while optimizing the average summation loss: $\frac{1}{T}\sum_{t=1}^T \hat{\epsilon}_t(\btheta^f,\btheta^c_t)$, \emph{i.e.}, compute the loss uniformly.
    \item Weighted MTL: Adapted from~\cite{murugesan2016adaptive}, learning a weighted summation of losses over different tasks: $\frac{1}{T}\sum_{t=1}^T \hat{\epsilon}_{\balpha_t}(\btheta^f,\btheta^c_t)$
    \item Adv.H: Adapted from \cite{liu2017adversarial} by using the same loss function while training with $\calH$-divergence as adversarial objective.
    \item Adv.W: Replace the adversarial loss of Adv.H by Wasserstein distance based adversarial training method.
    \item Multi-Obj.: Adapted from \cite{sener2018multi}, casting the multi-task learning problem as a multi-objective problem
    \item AMTNN: Adapted from~\cite{ijcai2019-478}, a gradient reversal layer with Wasserstein adversarial training method.
\end{itemize}

\subsection{Experiments on benchmark datasets}
We first evaluate the MTL algorithms on Digits dataset. In order to show the effectiveness of MTL methods when dealing with small amount of labelled instances, we follow the evaluation protocol of~\cite{ijcai2019-478} by randomly selecting $3k$, $6k$ and $8k$ instances of the training dataset and choose $1k$ dataset as validation set while testing one the full test set. For the SVHN dataset, we resize the images to $28\times 28 \times 1$, except for that, we do not apply any data-augmentation towards to digits dataset. A LeNet-5~\cite{lecun1998gradient} model is implemented as feature extractor and three 3-layer MLPs are deployed as task-specific classifiers, and extract the semantic feature from the classifier with size $128$. We adopt the Adam optimizer~\cite{kingma2014adam} for training the model from scratch. The model is trained for $50$ epochs while the initial learning rate is set by $1\times10^{-3}$ and is decayed $5\%$ for every $5$ epochs. The results are reported in Table~\ref{tab:digit}. 

For the computer vision applications, we then test the SMTL algorithm comparing with the baselines on PACS and Caltech datasests using the AlexNet~\cite{krizhevsky2012imagenet} as feature extractor. For investigating the performance when limited amount labelled instances are available, we evaluate the algorithms on PACS dataset randomly select $10\%$, $15\%$ and $20\%$ of the total dataset for training, respectively. Since this Office-Caltech dataset is relatively small, we only test the dataset by using $20\%$ of the total images to train the model. We use the pre-trained AlexNet provided by \emph{PyTorch}~\cite{paszke2019pytorch} while removing the last FC layers as feature extractor (out feature size 4096). On top of the feature extractor, we implement several MLPs as task-specific classifiers. The test results are reported in Table~\ref{Table_results_PACS} and Table~\ref{caltech}, respectively. After that, we then evaluate the algorithms on Office-31 and Office-Home dataset by randomly select $5\%$, $10\%$ and $20\%$ training samples with pre-trained ResNet-18 model of \emph{PyTorch} while removing the last FC layers as feature extractor (out feature size $512$). For these four vision benchmarks we follow the pre-processing and train/val/test protocol by~\cite{long2017learning,cao2018exploring,li2017deeper}. We adopt the Adam optimizer with initial learning rate $2\times 10^{-4}$ and decayed $5\%$ every $5$ epochs while totally training for $80$ epochs. For stable training, we also enable the \emph{weight-decay} in Adam optimizer to enforce a $L_2$ regularization. The test results are reported in Table~\ref{Results_office-31} and~\ref{results_office-home}, respectively. 
 \emph{For more details about the experimental implementations, please refer to the supplementary materials.}

From Table.~\ref{tab:digit}$\sim$~\ref{results_office-home}, we could observe that our proposed method could outperform the baselines and improve the benchmark performances with state-of-the-art performances. Particularly, we found that when there are only few of labelled instances (\emph{e.g.}~$5\%$ of the total instances), our method could have a large margin of improvements regarding the baselines. \emph{This confirms the effectiveness of our methods when dealing with limited data.}

\begin{figure}
    \centering
    \includegraphics[width=0.65\linewidth]{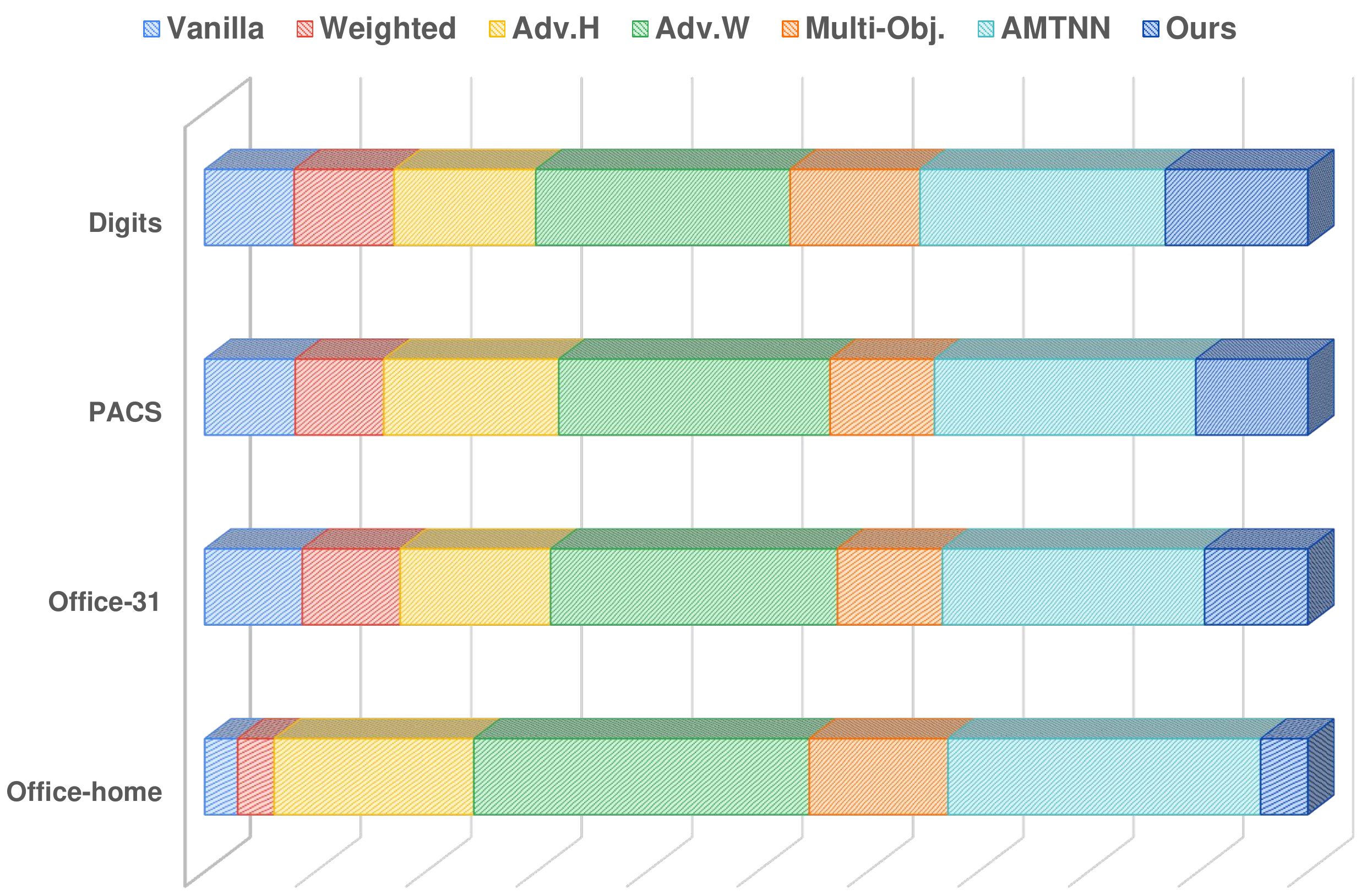}
    \caption{Relative time comparison (one training epoch) of the MTL algorithm on different benchmarks.}
    \label{time_comparision}
\end{figure}

\subsection{Further analysis}

\paragraph{Ablation studies}
In order to investigate the effectiveness of each component of our method, we conduct ablation studies (Table~\ref{ablation_studies}) of the proposed method on Office-31 dataset ($20\%$ of total instances) with four ablations, namely $1)$~\emph{Cls. only}: remove all of the re-weighting scheme, semantic matching and the convex optimization towards updating $\balpha$; $2)$~\emph{w.o.} re-weighting: removing the re-weighting scheme inside the label weighting loss; $3)$~\emph{w.o.} sem. matching: omitting the semantic matching; and $4)$~\emph{w.o.} cvx. opt.: omit the optimization procedure for updating $\balpha$, \emph{i.e,} Eq.~(\ref{convex_opt_alpha}). The results showed that the label re-weighting scheme is crucial for the algorithm. Besides, we also observe $-1.0\%$ drop when omitting the semantic matching procedure and $-0.5\%$ once we omit the convex optimization procedure for $\balpha$.

\paragraph{Time efficiency}
As our method doesn't rely on adversarial training, it has better time efficiency. 
We compare the time-efficiency of the MTL algorithms on Digits ($8k$), PACS ($20\%$), Office-31 ($20\%$) and Office-home ($20\%$) datasets, and report the time comparison of one training epoch in a relative percentage bar chart in Fig.~\ref{time_comparision}. The adversarial based training methods (\emph{Adv.H, Adv.W} and \emph{AMTNN}) take longer time for a training epoch, especially on the Office-home dataset. Take the improved performance (Table~\ref{tab:digit}$\sim$\ref{results_office-home}) into consideration. Our method could improve that benchmark performance while reduce the time needed for training. \emph{This also demonstrates the benefits of algorithm in terms of time-efficiency.}

\begin{figure}
    \centering
    \includegraphics[width=0.48\linewidth]{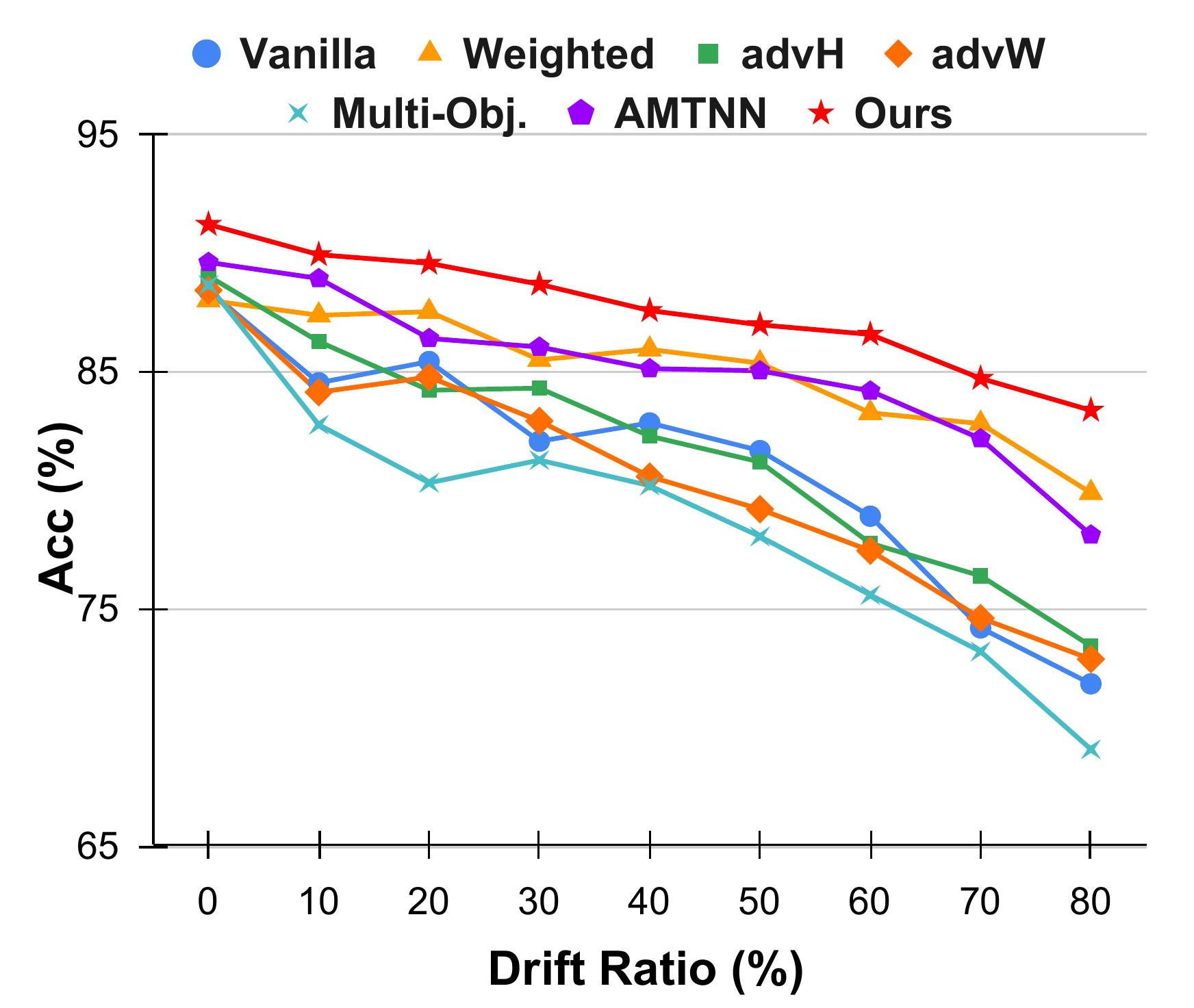}
    \includegraphics[width=0.48\linewidth]{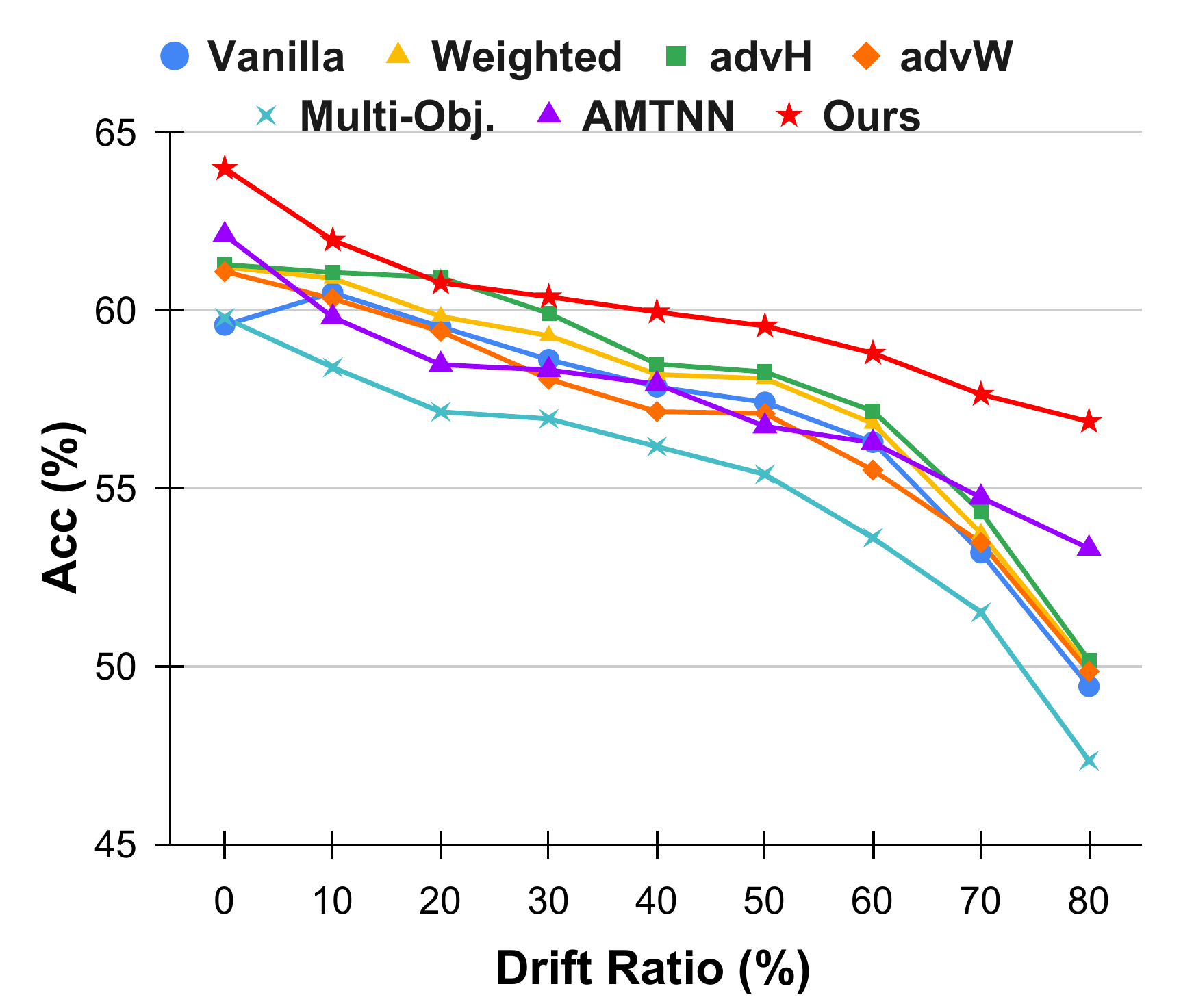}
    \caption{Performance comparison under label distribution drift scenario. \emph{Left}: Evaluations on Office-31 dataset with different drift ratio; \emph{Right}: Evaluations on Office-Home dataset with different drift ratio.}
    \label{fig:performance_compare_office31_home}
\end{figure}

\paragraph{Performance under label shift}
To confirm the effectiveness, we evaluate the MTL algorithms' performance under label shift situation where the label distribution drifts, \emph{i.e.}, the number of classes keeps the same with original one while some classes drift by a certain percentage for a specific task on Office-31 and Office-home dataset. The drift simulation is implemented as keeping all the classes within all the tasks while simulating a significant label distribution drift by randomly drop out some part of the instances of certain tasks. For Office-31 dataset, the task \emph{Amazon}'s class $1\sim10$, task \emph{Dslr}'s class $10\sim20$ and task \emph{Webcam}'s class $21\sim30$ are drifted with different ratios ($10\%\sim80\%$) while for Office-home dataset, we drift classes $1\sim16$ of \emph{Art}, classes $17\sim32$ of \emph{Clipart}, classes $33\sim48$ of \emph{Product}, and classes $49\sim64$ of \emph{Real World} with different ratios ($10\%\sim80\%$).

 We show the performance under label distribution drift ranging from $10\%\sim80\%$ on Office-31 dataset (left) and on Office-Home dataset (right) in Fig.~\ref{fig:performance_compare_office31_home}.
 As we could observe from Fig.~\ref{fig:performance_compare_office31_home}, when the label space drifts, all the algorithms drop off. Our algorithm could outperform the baselines with a large margin when label space shift. \emph{This demonstrates the benefits of our algorithm for handling label shift problems.} 
\section{Conclusion}
We propose to leverage the labeling information across different tasks in multi-task learning problems. 
We first theoretically analyze the generalization bound of multi-task learning based on the notion of Jensen-Shannon divergence, which provides new insights into the value of label information by exploiting the semantic conditional distribution in multi-task learning. Our theoretical results also lead to a concrete algorithm that jointly matches the semantic distribution and controls label distribution divergence. The empirical results demonstrates the effectiveness of our algorithm on improving the benchmark performance with better time efficiency and particularly show the benefits when label distribution shift.

\section*{Acknowledgement}
The authors would like to thank Changjian Shui for proofreading the manuscript as well as the constructive discussions, and also thank Shichun Yang and Jean-Philippe Mercier for the help with computational resources. This work has been supported by Natural Sciences and Engineering Research Council of Canada (NSERC). Fan Zhou is supported by the China Scholarship Council. Boyu Wang is supported by the NSERC, Discovery Grants Program.

\bibliography{reference}
\end{document}